\documentclass[letterpaper, 10 pt, conference]{ieeeconf}

\IEEEoverridecommandlockouts                % needed to use \thanks
\usepackage{graphicx}
\usepackage{float}
\usepackage{amsmath}
\usepackage{amsfonts}
\usepackage{amssymb}
\usepackage{booktabs}
\usepackage{multirow}
\usepackage{wrapfig}
\usepackage{xspace}
\usepackage{cite}                           % compresses [1],[2],[3] into [1]--[3]
\usepackage[hidelinks]{hyperref}
\hypersetup{
    pdftitle={DexTaG: Tactile-as-Guidance in Reinforcement Learning for Dexterous Manipulation},
    pdfauthor={Han Yang, Yian Wang, Yunlong Song, Zhenjia Xu, Chuang Gan},
    pdfkeywords={dexterous manipulation, reinforcement learning, tactile sensing}
}

\renewcommand{\paragraph}[1]{\par\vspace{0.5ex}\noindent\textbf{#1}\hspace{0.4em}\ignorespaces}

\newcommand{\methodname}{DexTaG\xspace}

\title{\LARGE \bf
DexTaG: \underline{T}actile-\underline{a}s-\underline{G}uidance in Reinforcement
Learning for\\
\underline{Dex}terous Manipulation
\\[0.25em]
{\normalfont\normalsize\url{https://dextag.github.io/}}
\par\vspace{-8pt}
}

\author{Han Yang$^{1}$, Yian Wang$^{1}$, Yunlong Song$^{2}$, Zhenjia Xu$^{2}$, and Chuang Gan$^{1}$%
\thanks{$^{1}$University of Massachusetts Amherst.}%
\thanks{$^{2}$Genesis AI.}%
}

\begin{document}

%===============================================================================
\makeatletter
\IEEEaftertitletext{%
\vspace{-10pt}
\begin{minipage}{\textwidth}
    \centering
    \def\@captype{figure}
    \setlength{\abovecaptionskip}{-2pt}
    \includegraphics[width=\linewidth]{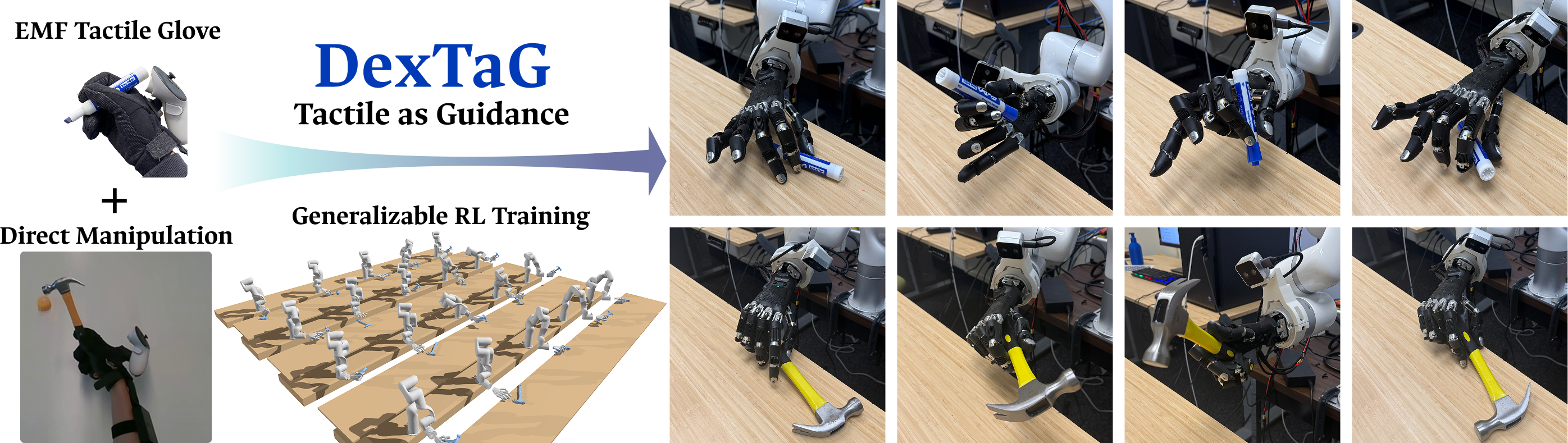}
    \caption{We present \methodname, a tactile-guided RL framework for dexterous manipulation that uses contact measurements from a data-collection glove. From data collection to policy training, the pipeline enables non-trivial tool-use tasks involving in-hand reorientation.}
    \label{fig:teaser}
\end{minipage}
\vspace{0.5\baselineskip}
}
\makeatother

\maketitle
\thispagestyle{empty}
\pagestyle{empty}

%===============================================================================
\begin{abstract}
Glove-based motion capture is emerging as a scalable approach to collecting dexterous-hand demonstration data. However, due to the kinematic gap between the human and robot hand, the recorded human motions cannot be executed directly on the robot, especially for contact-rich tool-use tasks involving in-hand reorientation. Prior work bridges this gap in simulation through reinforcement learning (RL) or trajectory optimization, but the human contact pattern is hard to preserve under such formulations, often producing unnatural manipulation and unstable functional grasps. These methods also train a separate policy or solve a separate optimization for each reference trajectory, which is inefficient. To solve these problems, we propose \emph{\methodname}, a tactile-guided RL framework for dexterous manipulation. During training, tactile signals captured by the glove guide policy search toward the measured human contact pattern, reducing reliance on precise reference geometry for contact supervision. To improve efficiency, we train a single generalizable retargeter jointly on all training trajectories of the same object. The retargeter is further distilled into a tactile-free student controller conditioned on the target object trajectory for real-world deployment. On marker-pen and hammer manipulation tasks, \methodname learns natural, contact-rich behaviors that baselines with distance-based contact heuristics fail to learn, generalizes to held-out trajectories of the same object and task, and outperforms single-trajectory baselines on OakInk2.
\end{abstract}

%===============================================================================
\section{Introduction}

Data remains a central bottleneck of robot manipulation, particularly for high-dimensional, contact-rich dexterous hands. Teleoperation~\cite{qin2023anyteleop, wang2024dexcap, handa2020dexpilot, zhang2025doglove, gao2025glovity} is limited by the human--robot kinematic gap and the lack of direct contact-force feedback, making tasks such as picking up a flat-lying tool and reorienting it in-hand difficult to demonstrate. Exoskeleton-style rigs~\cite{xu2025dexumi, fang2025dexop, zhu2026whed, zhu2026dexexo} restore some force feedback, but constrain the collector's workspace and finger dexterity and require hardware tailored to each robot.

Glove-based motion capture is emerging as a more scalable alternative. Wearing an EMF (electromagnetic-field) motion capture glove, the demonstrator manipulates objects directly with essentially bare-hand dexterity, while the glove records the hand pose. However, the human motions cannot be replayed on the robot due to the kinematic gap between the human and robot hand. Prior work bridges this gap in simulation by training a reinforcement learning (RL) policy~\cite{christen2022d, wang2023physhoi, zhang2024artigrasp, li2025maniptrans} or solving a trajectory optimization~\cite{liu2024parameterized, lakshmipathy2024kinematic, pan2025spider} that tracks the recorded hand--object trajectory on the robot. These methods have two drawbacks. First, hand--object pose references contain errors from vision-based object-pose estimation, which can make motion tracking conflict with contact acquisition. In our collected demonstrations, reference fingertips can remain separated from the object during pickup, particularly for thin objects such as a hammer handle. Tracking both hand and object poses can therefore discourage the adjustments needed to establish contact. Distance-based contact heuristics~\cite{li2025maniptrans, pan2025spider} inherit the same imperfect reference geometry, whereas glove tactile signals directly capture the measured human contact pattern. Second, these methods fit a separate policy or optimization for \emph{each} reference trajectory, which is computationally expensive as demonstration datasets grow.

To solve these problems, we propose \emph{\methodname}, an RL framework for \underline{Dex}terous manipulation using \underline{T}actile signals from the data-collection glove \underline{a}s \underline{G}uidance. Our pipeline has three features. \textbf{(i) Tactile-as-Guidance.} During training, tactile signals captured by the glove guide policy search toward the measured human contact pattern, reducing reliance on precise reference geometry for contact supervision. We incorporate the measured full-hand tactile readouts into the RL reward to guide contact acquisition and grasp configuration, rather than relying on contact inferred from reference geometry. \textbf{(ii) Efficient generalizable retargeting.} Rather than training a separate policy for each reference trajectory, we train a \emph{single} generalizable retargeter jointly on all training trajectories of the same object. A long-horizon window of future reference frames helps the policy generalize to held-out references within the same task without per-trajectory retraining. \textbf{(iii) Distillation for deployment.} The retargeter is distilled into a tactile-free student controller conditioned on a target object trajectory and using visual--proprioceptive feedback, making it deployable in the real world without tactile hardware on the robot.

We evaluate \methodname on two tool-use tasks---marker-pen and hammer---both of which require non-trivial pickup of a flat-lying tool and in-hand reorientation toward a functional grasp. Ablations on the tactile-guidance rewards show that they enable grasp acquisition from noisy pose references where training without these rewards fails, while also encouraging more human-like manipulation and contact patterns. Real-world experiments further demonstrate execution with the distilled tactile-free controller. In simulation, increasing the number of training references generally improves tracking performance on both training and held-out trajectories for the retargeter and controller. On scaling comparisons against ManipTrans on OakInk2, our pipeline maintains performance as the number of reference trajectories increases, while the baseline variants degrade.

Our contributions are:

\begin{enumerate}
    \setlength{\itemsep}{1pt}
    \setlength{\parskip}{1pt}
    \setlength{\topsep}{1pt}
    \item \textbf{Tactile-as-Guidance}, a reward formulation that uses measured glove tactile readouts to guide contact acquisition and human-like grasp formation alongside hand--object motion tracking.
    \item \textbf{An efficient RL pipeline for dexterous manipulation} that trains a single generalizable retargeter across training trajectories of the same object, supports held-out references within the same task, and distills the retargeter into a real-world-deployable tactile-free student controller.
    \item \textbf{Empirical evidence of efficiency and generalization}: across two contact-rich tool-use tasks and an OakInk2 scaling comparison, \methodname maintains performance as trajectories are added, generalizes to held-out references, and outperforms per-trajectory baselines.
\end{enumerate}

%===============================================================================
\section{Related Work}

\paragraph{Tracking Reference Hand--Object Trajectories.}
Many dexterous manipulation methods fit a separate policy or optimization to each reference trajectory~\cite{li2025maniptrans, mandi2025dexmachina, li2026physgraph, pan2025spider, liu2024parameterized, reda2023physics, lakshmipathy2024kinematic, yang2025physics}. Recent work explores generalizable tracking: DexTrack~\cite{liu2025dextrack} combines per-trajectory RL demonstration mining with shared-controller training; HOT~\cite{wang2025learning} distills skill- or object-specific trackers learned from synthetic references; and DexSynRefine~\cite{lee2026dexsynrefine} uses a generative prior fitted to a small demonstration set. Motion-capture datasets provide reference trajectories~\cite{zhan2024oakink2, taheri2020grab}. We train a single tactile-guided retargeter directly across glove-captured references of the same object, without per-trajectory policy optimization, and evaluate generalization to held-out references within the same task.

\paragraph{Demonstration Data for Dexterous Manipulation.}
Demonstrations can be collected through vision- or glove-based teleoperation~\cite{yang2024ace, ding2025bunny, tang2026towards, cui2025end, du2025mile, jia2026feel} and direct-manipulation exoskeletons~\cite{xu2025dexumi, fang2025dexop, si2025exostart, zhu2026whed, zhu2026dexexo}, with trade-offs in embodiment mismatch, force feedback, and operator dexterity. Glove-based motion capture instead records direct human manipulation for VLA pretraining~\cite{zheng2026egoscale, qiu2025humanoid} or downstream policy learning~\cite{lee2026dexsynrefine}. Other approaches synthesize references using rules~\cite{wang2025learning}, simulation-based optimization~\cite{yang2025physics}, or demonstration augmentation~\cite{bai2026far, jiang2025dexmimicgen}. We use glove-captured motion and measured contact signals to supervise robot learning.

\paragraph{Tactile Sensing in Manipulation.}
Prior work uses tactile feedback as a policy input on grippers~\cite{akinola2025tacsl, huang2025vt} and dexterous hands~\cite{ye2026visual, lin2025learning}, develops tactile simulation for transfer~\cite{su2026tacmap, zhao2026closing}, and integrates tactile sensors into demonstration hardware~\cite{du2025mile, fang2025dexop}. Other methods use tactile targets to pretrain visual--tactile representations for control~\cite{ye2026visual, liu2025vtdexmanip, sun2025vtao}. Closely related, PTLD~\cite{chen2026ptld} uses robot tactile data to train a contact-state estimator from proprioception. We instead use recorded \emph{human} tactile signals as direct RL reward guidance for contact acquisition and grasp configuration. This supervision differs from online robot tactile feedback: the retargeter observes simulated tactile signals, while the distilled student requires no tactile sensing at deployment.

%===============================================================================

\section{Method}

\begin{figure*}[t]
    \centering
    \includegraphics[width=\linewidth]{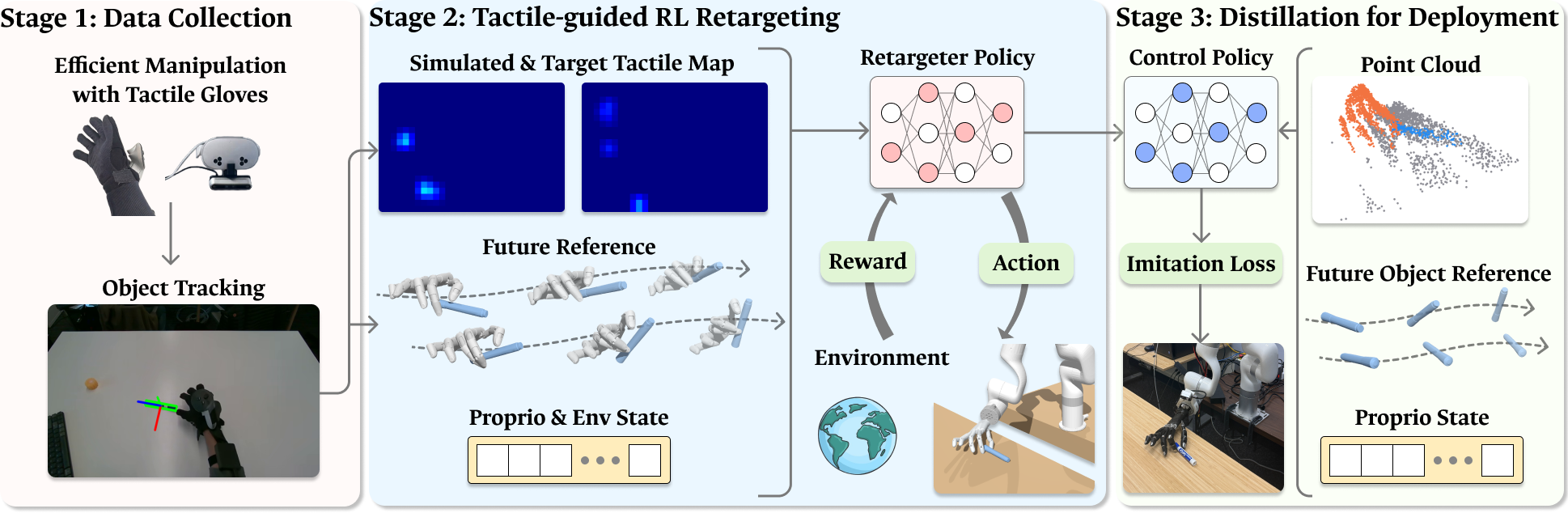}
    \vspace{-20pt}
    
    \caption{\textbf{Overview of \methodname.} The pipeline consists of three stages: (i) collecting hand--object demonstrations with tactile gloves, (ii) using measured tactile references to guide RL retargeting, and (iii) distilling the retargeter into a tactile-free controller for real-world deployment.}
    
    \label{fig:pipeline}
    \vspace{-12pt}
\end{figure*}

\methodname consists of three stages (Fig.~\ref{fig:pipeline}). \textbf{(i) Tactile demonstration collection and preprocessing} (Sec.~\ref{sec:data_collection_pipeline}): the demonstrator manipulates objects directly while wearing a tactile glove, producing hand--object trajectories paired with tactile maps. \textbf{(ii) Tactile-guided RL retargeting} (Sec.~\ref{sec:generalizable_retargeter}): measured contact signals guide policy search through the RL reward, helping the robot acquire grasps and reproduce human contact patterns alongside motion tracking. Rather than training a separate policy for each reference trajectory, we train a single generalizable retargeter $\pi_{\text{rtgt}}$ jointly on all training trajectories of the same object. The trained policy generalizes to held-out references within the same task, allowing newly collected demonstrations to be retargeted directly without per-trajectory retraining. \textbf{(iii) Distillation for deployment} (Sec.~\ref{sec:generalizable_controller}): $\pi_{\text{rtgt}}$ is distilled into a tactile-free student controller $\pi_{\text{ctrl}}$ whose reference input contains only a target object trajectory. Using visual--proprioceptive feedback, the student is deployable in the real world. Its object-trajectory-only interface also permits targets to be specified through rule-based procedures.

\subsection{Tactile Demonstration Collection and Preprocessing}
\label{sec:data_collection_pipeline}

\begin{figure}[H]
  \centering
  \setlength{\abovecaptionskip}{-2pt}
  \begin{minipage}[t]{0.48\linewidth}
    \vspace{0pt}\centering
    \includegraphics[width=\linewidth]{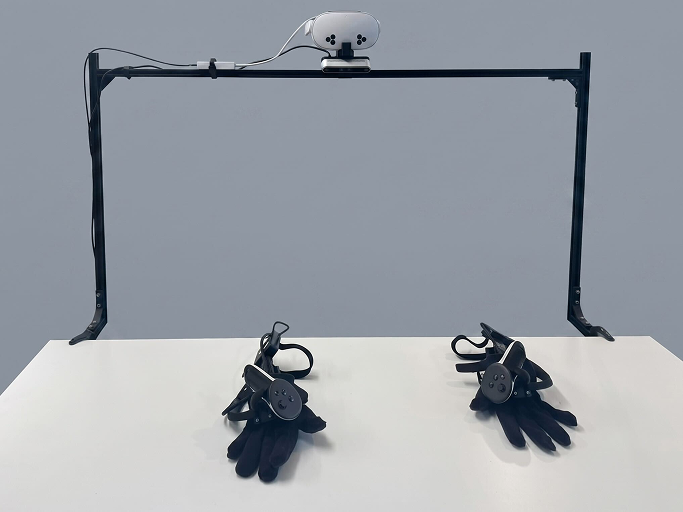}
    {\footnotesize\caption{Data collection hardware: a pair of WUJI tactile gloves and an overhead camera for object pose estimation.}\label{fig:data_collection_hardware}}
  \end{minipage}\hfill
  \begin{minipage}[t]{0.48\linewidth}
    \vspace{0pt}\centering
    \includegraphics[width=\linewidth]{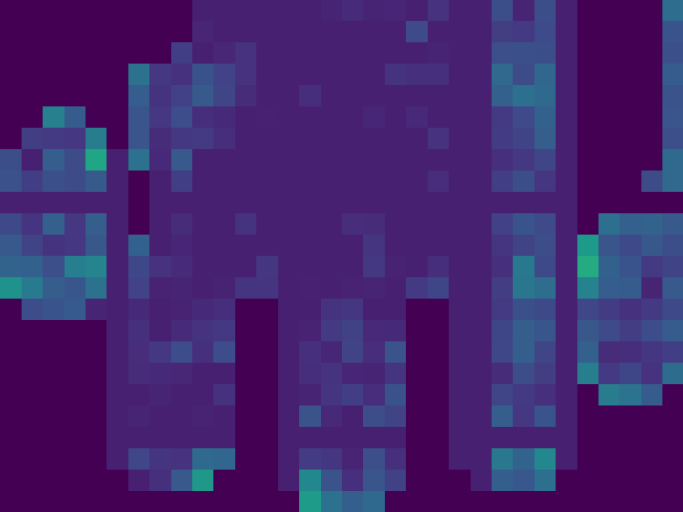}
    {\footnotesize\caption{Raw $24 \times 32$ tactile map readout from the glove during manipulation.}\label{fig:tactile_glove_readout}}
  \end{minipage}
  \vspace{-2mm}
\end{figure}

\paragraph{Data acquisition.}
Our data-collection setup, shown in Fig.~\ref{fig:data_collection_hardware}, lets the demonstrator manipulate objects directly while wearing a WUJI tactile glove. A pressure-sensor array covering the fingers and palm provides a $24 \times 32$ tactile map (Fig.~\ref{fig:tactile_glove_readout}), while five EMF modules record fingertip poses relative to the wrist. A Meta Quest controller mounted on the back of the glove tracks its global motion, and a calibrated overhead RealSense camera records RGB-D observations for object-pose estimation. 

\vspace{-2mm}

\paragraph{Tactile background calibration.}
The glove can produce nonzero tactile readings even without object contact, due to sensor offsets and self-contact as the fingers curl. To estimate this background, the demonstrator holds several representative hand poses encountered during manipulation, such as a half-closed fist, without an object in hand. We record a tactile map for each pose and take the pixelwise maximum across these maps to obtain a single fixed background map. The same background map is subtracted from every frame of the recorded tactile data to suppress non-object-contact responses before the signals are used for contact guidance.

\paragraph{Hand--object reference construction.}
We obtain target robot hand joint angles through kinematic retargeting of the recorded fingertip poses and recover the wrist pose using a fixed controller-to-wrist transformation. Object poses are estimated from the calibrated RGB-D recordings using FoundationPose~\cite{foundationposewen2024} and a reconstructed object mesh. The resulting reference trajectories contain synchronized robot hand joint angles, wrist poses, object poses, and background-corrected tactile maps.

\begin{figure}[H]
    \centering
    \setlength{\abovecaptionskip}{-2pt}
    \includegraphics[width=\linewidth]{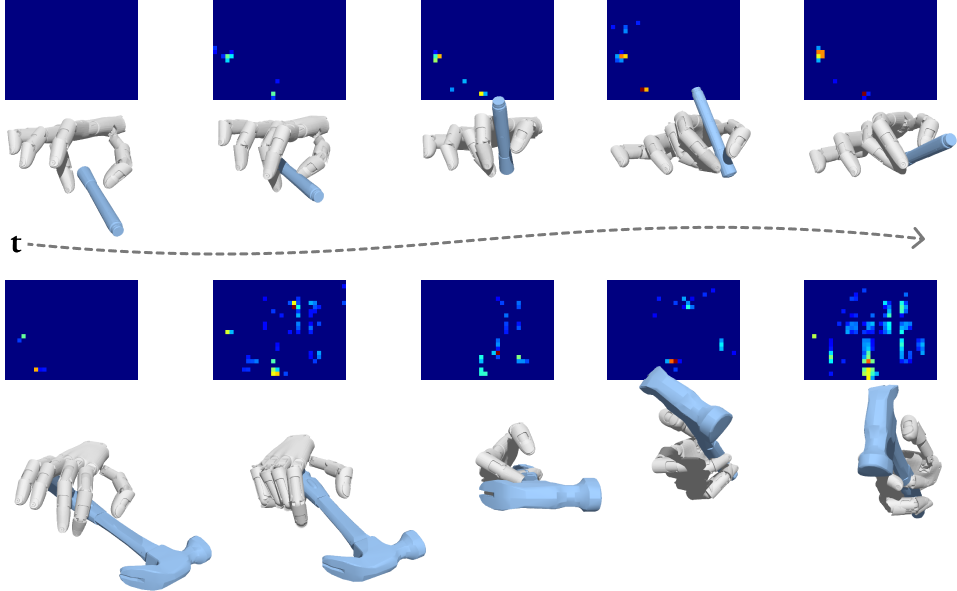}
    {\footnotesize\caption{Example collected trajectories with per-timestep tactile maps for marker-pen (top) and hammer (bottom) manipulation.}\label{fig:data_visualization}}
\end{figure}

We collect $100$ demonstrations for marker-pen manipulation and $140$ for hammer manipulation. Each demonstration includes approaching and picking up the tool, reorienting it in-hand toward a functional grasp, performing a functional motion, and placing it back. Figure~\ref{fig:data_visualization} shows example trajectories and their corresponding tactile readouts.

\vspace{-10pt}

\subsection{Tactile-Guided RL Retargeting}
\label{sec:generalizable_retargeter}

We formulate retargeting as tracking reference hand--object motion while using measured tactile signals to guide contact acquisition and grasp configuration. Let $\boldsymbol{\tau}=(\boldsymbol{\xi}_0,\ldots,\boldsymbol{\xi}_T)$ denote a motion reference, where $\boldsymbol{\xi}_t$ contains the target hand joint state, wrist and fingertip poses, and object motion at timestep $t$. Each frame is paired with a reference tactile map $\mathbf{t}^{\text{ref}}_t$, kept separate from the motion reference. We first describe how tactile signals guide policy search, then the shared-policy training setup.

\subsubsection{Tactile-as-Guidance Rewards}
\label{sec:tactile_reward_design}

The glove's tactile readouts provide measured contact information that is difficult to infer reliably from noisy hand--object pose references. We use these signals to shape policy search alongside motion tracking through two rewards: fingertip proximity encourages contact acquisition, while tactile-map similarity guides the spatial contact pattern across the hand.

\paragraph{Simulated tactile correspondence.}
We implement an SDF-based full-hand tactile sensor in Genesis~\cite{Genesis}, with approximately $700$ sample points on the robot hand's palm-facing surfaces. At each point, simulated normal-force magnitude is estimated from object penetration depth using a stiffness of $2000$~N/m. We manually specify correspondences between robot-hand regions and regions of the glove's $24 \times 32$ layout, then assign sample points to pixels within these regions. Each pixel reports the mean force magnitude of its assigned points. This establishes a spatial correspondence. Let $\mathbf{t}^{\text{sim}}_t,\mathbf{t}^{\text{ref}}_t \in \mathbb{R}^{P}$ denote the flattened simulated and reference tactile maps, with $P=768$.

\paragraph{Tactile-gated fingertip proximity.}
This term supervises \emph{which} fingertips should be in contact, directly from the reference tactile readout. It encourages contact for fingertips that the reference marks as touching the object, leaving the remaining fingers unconstrained, and provides a dense learning signal that is especially important early in training, before the policy has discovered the contact pattern. Formally, for each fingertip $i \in \{1, \dots, 5\}$, let $\mathcal{P}_i \subset \{1,\dots,P\}$ be its assigned pixel set under the glove layout and $d_i$ the signed distance from the fingertip to the object surface. We define a per-finger contact gate and the proximity reward as follows:

\begingroup
\setlength{\abovedisplayskip}{-2pt}
\setlength{\belowdisplayskip}{4pt}
\setlength{\abovedisplayshortskip}{-4pt}
\setlength{\belowdisplayshortskip}{0pt}
\begin{equation}
\begin{aligned}
g_i \;&=\; \mathbf{1}\!\left[\max_{p \in \mathcal{P}_i} \mathbf{t}^{\text{ref}}_{t,p} \,>\, 0\right], \\[2pt]
r^{\text{prox}}_t \;&=\; \sum_{i=1}^{5} g_i \,\cdot\, \exp\!\left(-k_{\text{prox}}\,\max(d_i,\,0)\right).
\end{aligned}
\label{eq:tactile_proximity_reward}
\end{equation}
\endgroup

\paragraph{Tactile-map similarity.}
Beyond enforcing \emph{which} fingertips contact the object, this term guides the contact pattern on the palm and finger pads as well as the fingertips. Since the simulated and reference maps come from different sensing processes, their magnitudes are not directly comparable.

To reduce this scale difference, we normalize each map independently by its own peak and apply gamma correction. For either input map $\mathbf{x}$, the transformation is
\begin{equation}
\mathcal{N}(\mathbf{x})_p = \sqrt{\operatorname{clip}\!\left(
\frac{x_p}{\max(1,\max_q x_q)},\,0,\,1\right)}.
\label{eq:tactile_peak_normalization}
\end{equation}
The denominator floor prevents division by zero and excessive amplification of weak signals. The square root ($\gamma=0.5$) compresses dynamic range, increasing the relative contribution of weaker contacts. The transformed maps lie in $[0,1]$ and are denoted by $\widehat{\mathbf{t}}^{\text{sim}}_t=\mathcal{N}(\mathbf{t}^{\text{sim}}_t)$ and $\widehat{\mathbf{t}}^{\text{ref}}_t=\mathcal{N}(\mathbf{t}^{\text{ref}}_t)$.

We then apply a $3\times3$ Gaussian smoothing kernel $G_\sigma$ with $\sigma=1.0$ to both maps to tolerate small spatial offsets between human and robot contact regions. The discrepancy between the processed maps is converted into an exponentially decaying similarity reward:
\begin{equation}
\begin{aligned}
E^{\text{tac}}_t &= \mathcal{L}_1\!\left(
G_\sigma\star\widehat{\mathbf{t}}^{\text{sim}}_t,
G_\sigma\star\widehat{\mathbf{t}}^{\text{ref}}_t\right), \\[2pt]
r^{\text{tac}}_t &= \exp\!\left(-k_{\text{tac}}E^{\text{tac}}_t\right).
\end{aligned}
\label{eq:tactile_similarity_reward}
\end{equation}
Here, $\star$ denotes convolution and $\mathcal{L}_1$ is the $L_1$ loss. We set the reward decay parameters to $k_{\text{prox}}=50$ and $k_{\text{tac}}=4.0$. Together, normalization, gamma correction, and spatial smoothing let the reward compare contact patterns across the two sensing domains without requiring agreement in absolute force magnitudes.

\subsubsection{Generalizable Retargeter Training}
\label{sec:algorithm_overview}

\paragraph{Policy observations and optimization.}
The retargeter actor observes the motion reference, current simulated robot--object state $\mathbf{p}_t$, current simulated tactile map, and a future-reference window:
\begin{equation}
\mathbf{o}_t=\bigl(\boldsymbol{\xi}_t,\,\mathbf{p}_t,\,\mathbf{t}^{\text{sim}}_t,\,\boldsymbol{\xi}_{t:t+K}\bigr).
\label{eq:retargeter_actor_observation}
\end{equation}
The critic additionally receives the reference tactile map and the past $8$ steps of proprioception--action history; neither is an actor input. Recorded human tactile therefore provides reward supervision and privileged critic information, while the actor observes the robot's current simulated tactile response. The actor outputs $27$-dimensional arm--hand joint-position commands tracked by PD controllers. We train with PPO~\cite{schulman2017proximal}, combining the two tactile rewards with wrist, fingertip, hand-joint, and object tracking rewards, plus stability and contact penalties.

\paragraph{Training across demonstrations.}
Rather than training a separate policy for each reference trajectory, we train a single generalizable retargeter $\pi_{\text{rtgt}}$ jointly on all training trajectories of the same object, with the aim of generalizing to held-out references within the same task. The motivation is that demonstrations of the same object and task share manipulation substructure, such as lift motions and an eventual functional grasp, while differing in their approach, pickup, and in-hand reorientation paths. A shared policy can therefore reuse these common behaviors across references rather than learning an isolated solution for each trajectory. To support this transfer, the actor observes a future-reference window of length $K=32$, subsampled at stride $2$ to give $16$ frames and encoded by a temporal convolutional network. This temporal context reveals the direction and timing of upcoming motion, allowing the policy to anticipate subsequent manipulation stages rather than react only to the current reference frame. Trajectory-level SE(2) and scale augmentations further expose the policy to variations in placement, orientation, and scale.

\begin{figure*}[t]
    \centering
    \setlength{\abovecaptionskip}{-2pt}
    \includegraphics[width=\linewidth]{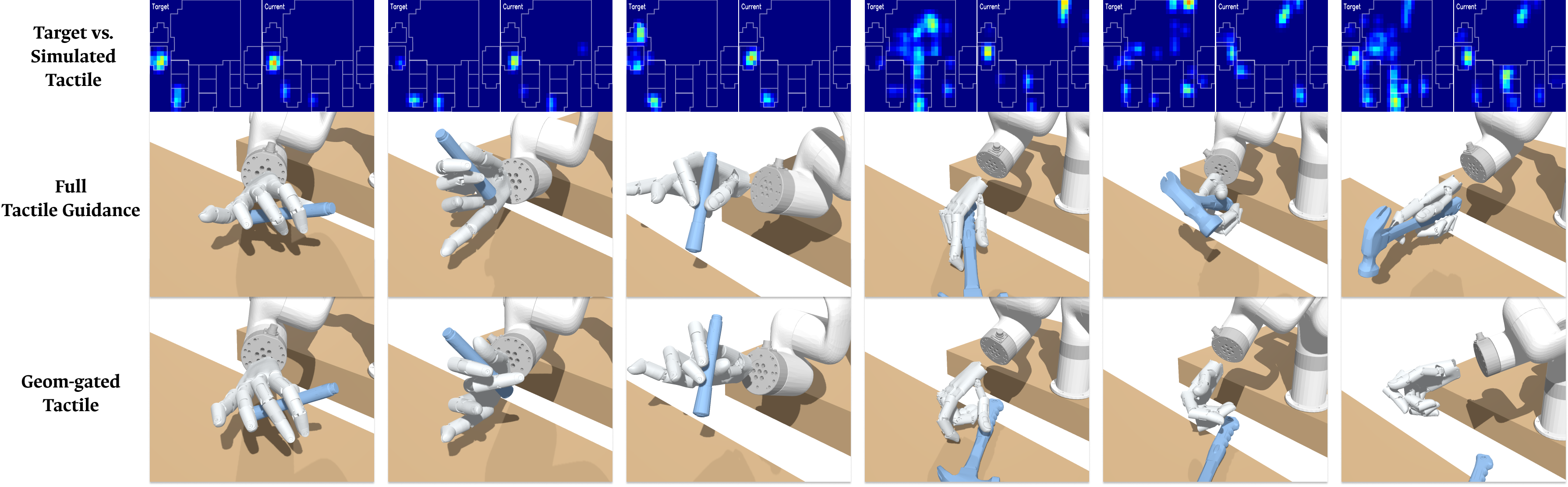}
    \caption{\textbf{Qualitative Comparison.} \emph{First row:} the reference glove tactile map vs. the simulated tactile map at each timestep. \emph{Second and third rows:} policy rollouts of \methodname and Geom-gated, respectively, on marker-pen and hammer manipulation.}
    \label{fig:qualitative}
    \vspace{-6pt}
\end{figure*}

\subsection{Distillation for a Deployable Controller}
\label{sec:generalizable_controller}

We distill $\pi_{\text{rtgt}}$ into a tactile-free student controller $\pi_{\text{ctrl}}$ via behavior cloning with DAgger~\cite{ross2011reduction}. A key design choice is that the student's reference input contains only a target \emph{object} trajectory---no reference hand pose. The goal is to follow both training and unseen target object trajectories within the same object and task setting. This choice reflects what is practical to specify at deployment: object trajectories can be captured from video or directly synthesized by simple rule-based procedures (e.g., a parametric writing path for a marker pen), whereas dense reference hand-and-finger trajectories are difficult to acquire outside a tactile-glove collection rig.

The student observes arm--hand joint angles, a window of future target object poses, a wrist-camera point cloud encoded by PointNet~\cite{qi2017pointnet}, and a short proprioception--action history. The point cloud replaces the teacher's privileged object pose. The same trajectory-level SE(2) and scale augmentations are applied during distillation to support tracking across variations in the target trajectory.

Tactile-free deployment is deliberate: the WUJI hand used in our experiments has no tactile sensors, and distillation avoids requiring real tactile observations to match the simulated sensor model. Human tactile signals thus guide behavior learning without imposing a tactile-sensing requirement on the deployed robot. The controller remains closed-loop through visual and proprioceptive feedback, but does not provide direct tactile feedback for detecting slip or unexpected contact.
%===============================================================================

\section{Experimental Results}

\paragraph{Experimental Setup.}
We evaluate \methodname with a 20-DoF WUJI right hand on a 7-DoF xArm: marker-pen and hammer tasks in simulation and the real world, and OakInk2~\cite{zhan2024oakink2} for baseline comparison.
Our simulation metric, \emph{Trajectory Completion}, is the fraction of a reference tracked before termination.
For our retargeter, Euclidean fingertip errors are limited to approximately $(57,64,71,86,86)$~mm, from thumb to little finger. Object-position/orientation limits are approximately $58$~mm/$43.7^\circ$, ignoring marker-pen axial rotation. Any exceeded limit or nonzero arm-link contact force terminates the episode.
The table-force limit decreases linearly through $500/100/50$~N at $0/10{,}000/20{,}000$ training steps, then stays fixed.
Sanity limits are $100$~m/s for wrist/object linear speed and $200$~rad/s for wrist/object angular speed or mean absolute hand-joint velocity.
We design our experiments to answer the following questions:

\begin{itemize}
    \setlength{\itemsep}{0pt}
    \setlength{\parskip}{0pt}
    \setlength{\topsep}{0pt}
    \item[\textbf{A.}] How do tactile rewards compare with no tactile rewards or geometry-derived contact gates?
    \item[\textbf{B.}] How does training-set size affect retargeter and controller performance on training and held-out references?
    \item[\textbf{C.}] How does our multi-trajectory fitting on OakInk2 compare with ManipTrans?
    \item[\textbf{D.}] Can the tactile-free controller execute the learned behaviors on real hardware?
\end{itemize}

\subsection{Tactile Reward Ablation}
\label{sec:exp_tactile_ablation}

\paragraph{Setup.}
We compare four alternatives with the same model, actor observations, and training schedule. \emph{Prox-only} retains tactile-gated fingertip proximity, while \emph{Sim-only} retains tactile-map similarity. \emph{Geom-gated} uses the same fingertip-proximity reward as Prox-only, but replaces the measured tactile gate with a binary contact label obtained by thresholding the fingertip--object distance in the motion-capture reference at $20$~mm (following ManipTrans~\cite{li2025maniptrans}). Finally, \emph{w/o tactile} removes both tactile-guidance rewards. All variants retain simulated tactile observations; Geom-gated and w/o tactile also omit the reference tactile map from the critic. For each variant and task, we train $\pi_{\text{rtgt}}$ with five random seeds and report mean $\pm$ standard deviation of Trajectory Completion and average per-step tracking rewards. \emph{Obj.} combines object-position and rotation rewards; \emph{Hand} combines wrist-position, wrist-rotation, and hand-joint rewards.

\begin{table}[t]
    \centering
    \scriptsize
    \setlength{\tabcolsep}{3pt}
    \setlength{\abovecaptionskip}{-2pt}
    \caption{\textbf{Tactile reward ablation.} Trajectory Completion (\%) and average per-step tracking rewards: mean $\pm$ standard deviation over five training seeds.}
    \label{tab:exp3_ablation}
    \begin{tabular*}{\linewidth}{@{\extracolsep{\fill}}lccc@{}}
        \toprule
        Method & Comp. & Obj. & Hand \\
        \midrule
        \multicolumn{4}{l}{\emph{Marker pen}} \\
        Ours & $\mathbf{87.5}\!\pm\!4.0$ & $1.127\!\pm\!0.033$ & $1.856\!\pm\!0.182$ \\
        Prox-only & $64.5\!\pm\!6.1$ & $1.106\!\pm\!0.064$ & $1.823\!\pm\!0.086$ \\
        Sim-only & $61.5\!\pm\!13.8$ & $1.100\!\pm\!0.097$ & $1.771\!\pm\!0.136$ \\
        Geom-gated & $48.6\!\pm\!24.0$ & $1.161\!\pm\!0.138$ & $1.983\!\pm\!0.315$ \\
        w/o tactile & $24.7\!\pm\!2.9$ & $1.299\!\pm\!0.020$ & $2.304\!\pm\!0.012$ \\
        \midrule
        \multicolumn{4}{l}{\emph{Hammer}} \\
        Ours & $\mathbf{76.8}\!\pm\!27.7$ & $1.158\!\pm\!0.052$ & $2.048\!\pm\!0.098$ \\
        Prox-only & $62.2\!\pm\!31.3$ & $1.139\!\pm\!0.087$ & $2.038\!\pm\!0.184$ \\
        Sim-only & $29.6\!\pm\!3.6$ & $1.217\!\pm\!0.024$ & $2.214\!\pm\!0.065$ \\
        Geom-gated & $29.9\!\pm\!5.5$ & $1.229\!\pm\!0.015$ & $2.195\!\pm\!0.059$ \\
        w/o tactile & $27.6\!\pm\!0.1$ & $1.237\!\pm\!0.007$ & $2.233\!\pm\!0.007$ \\
        \bottomrule
    \end{tabular*}
    \par\vspace{-16pt}
\end{table}

\begin{table*}[t]
    \centering
    \scriptsize
    \setlength{\tabcolsep}{2pt}
    \setlength{\abovecaptionskip}{-2pt}
    \begin{minipage}[t]{0.22\linewidth}
        \centering
        \scriptsize
        \setlength{\tabcolsep}{2pt}
        \caption{\textbf{Grasp-naturalness User Study} on marker-pen rollouts.}
        \label{tab:grasp_naturalness}
        \resizebox{\linewidth}{!}{%
        \begin{tabular}{@{}lcc@{}}
            \toprule
            Method & Votes & Preference (\%) \\
            \midrule
            \methodname & \textbf{92} & \textbf{55.8} \\
            w/o tactile & 40 & 24.2 \\
            Geom-gated & 33 & 20.0 \\
            \bottomrule
        \end{tabular}%
        }
    \end{minipage}\hfill
    \begin{minipage}[t]{0.40\linewidth}
        \centering
        \caption{\textbf{Generalizable retargeter $\pi_{\text{rtgt}}$.} Trajectory Completion (\%) on train / eval references: mean $\pm$ standard deviation over five random splits per ratio.}
        \label{tab:exp1_retargeter}
        \scriptsize
        \setlength{\tabcolsep}{2pt}
        \resizebox{\linewidth}{!}{%
        \begin{tabular}{@{}lcc@{}}
            \toprule
            Train:Eval & Marker Pen & Hammer \\
            \midrule
            1:9 & $55.1\!\pm\!20.9\;/\;43.8\!\pm\!13.1$ & $55.6\!\pm\!15.8\;/\;36.1\!\pm\!13.8$ \\
            5:5 & $75.7\!\pm\!4.9\;/\;69.9\!\pm\!6.3$ & $73.7\!\pm\!4.9\;/\;68.8\!\pm\!3.9$ \\
            9:1 & $73.3\!\pm\!5.0\;/\;75.4\!\pm\!5.0$ & $80.1\!\pm\!5.2\;/\;78.8\!\pm\!3.7$ \\
            \bottomrule
        \end{tabular}%
        }
    \end{minipage}\hfill
    \begin{minipage}[t]{0.32\linewidth}
        \centering
        \caption{\textbf{Generalizable controller $\pi_{\text{ctrl}}$.} Trajectory Completion (\%) on train / eval references after distillation, using one split per ratio.}
        \label{tab:exp1_controller}
        \resizebox{0.9\linewidth}{!}{%
        \begin{tabular}{@{}lccc@{}}
            \toprule
            Train:Eval & Marker Pen & Hammer & Avg. \\
            \midrule
            1:9 & 57.9 / 27.1 & 53.8 / 27.5 & 55.9 / 27.3 \\
            5:5 & 70.5 / 52.6 & 70.6 / 60.8 & 70.5 / 56.7 \\
            9:1 & 68.1 / 55.6 & 80.0 / 75.2 & 74.1 / 65.4 \\
            \bottomrule
        \end{tabular}%
        }
    \end{minipage}
    \par\vspace{-10pt}
\end{table*}

\paragraph{Results.}
Table~\ref{tab:exp3_ablation} shows complementary benefits from the two tactile rewards. Proximity-only increases mean completion over removing both rewards from $24.7\%$ to $64.5\%$ on marker pen and from $27.6\%$ to $62.2\%$ on hammer. Adding tactile-map similarity further increases completion to $87.5\%$ and $76.8\%$, respectively, supporting the combination of proximity-based contact acquisition and spatial contact-pattern guidance. However, hammer grasp acquisition remains seed-sensitive: one of the five full-method runs fails to learn pickup and achieves only approximately $27\%$ completion, contributing to the large standard deviation.

Measured contact gates also outperform geometry-derived gates on average: Prox-only achieves $64.5\%$ versus $48.6\%$ on marker pen and $62.2\%$ versus $29.9\%$ on hammer. Since the proximity-reward form is unchanged, this supports measured contact supervision over geometry-derived labels on the collected references. Fig.~\ref{fig:qualitative} shows representative rollouts: Geom-gated picks up the marker pen but forms an unnatural grasp, and fails to pick up the hammer. Average tracking rewards must be interpreted cautiously: policies that fail early can score higher by tracking only the easier approach phase.

\paragraph{Grasp naturalness.}
We further conduct a user study to assess grasp naturalness beyond tracking rewards. We focus on marker pen because successful pickups yield diverse grasp configurations, whereas hammer variants mainly differ in grasp acquisition.
Eleven respondents compared marker-pen rollouts of \methodname, w/o tactile, and Geom-gated. We randomly selected $15$ reference trajectories on which all methods reached the midpoint and showed palm-view midpoint renders in randomized, anonymized order. Respondents answered: \emph{``In each group, pick the most natural / most human-like grasp.''} \methodname received $55.8\%$ of $165$ votes (Table~\ref{tab:grasp_naturalness}), indicating a preference for its grasps naturalness.

\subsection{Retargeter and Controller Generalization}
\label{sec:exp_generalization}

\paragraph{Setup.}
For each of the marker pen and hammer tasks, we randomly split the trajectories into train/eval sets at ratios $\{1{:}9,\,5{:}5,\,9{:}1\}$, train one teacher $\pi_{\text{rtgt}}$ on the training trajectories, and distill it into a student controller $\pi_{\text{ctrl}}$.
We report Trajectory Completion on both training and held-out eval trajectories of the same object and task. For the retargeter, we repeat training with five independent random train/eval splits per ratio and report mean $\pm$ standard deviation to capture variability in RL training and demonstration selection.
Note that $\pi_{\text{rtgt}}$ is terminated on hand- or object-pose mismatch, while $\pi_{\text{ctrl}}$, which receives no reference hand pose, is terminated on object-pose mismatch only.

\paragraph{Results.}
Tables~\ref{tab:exp1_retargeter} and~\ref{tab:exp1_controller} report Trajectory Completion for the retargeter and controller. Two trends stand out. First, \emph{more training trajectories generally improve both training and held-out performance}, despite requiring one policy to fit more diverse trajectories. Marker-pen training performance largely plateaus after the 5:5 split, with a slight decrease at 9:1, but its held-out performance continues to improve. Hammer benefits from additional references on both training and evaluation trajectories. The five-split retargeter results show lower variability in held-out performance as the training set grows. Second, \emph{the train/eval gap narrows for both the retargeter and controller}. Together with improved held-out performance, this suggests that additional demonstrations help the policy learn behaviors that transfer across references, rather than simply fit individual training trajectories. This generalization is retained after distillation: the tactile-free controller follows held-out collected object trajectories, without reference hand poses or per-trajectory retraining.

\begin{table*}[t]
    \centering
    \setlength{\abovecaptionskip}{-2pt}
\begin{minipage}[t]{0.53\linewidth}
    \centering
    \scriptsize
    \setlength{\tabcolsep}{3pt}
    \caption{\textbf{Scaling on OakInk2.} Trajectory Completion (\%) with $n$ references per policy. Avg. is trajectory-weighted; $\Delta$ is the relative change from $n{=}1$ to $n{=}\text{all}$.}
    \label{tab:exp2_oakink2}
\begin{tabular*}{\linewidth}{@{\extracolsep{\fill}}llccccc@{}}
            \toprule
            Method & Object & $n{=}1$ & $n{=}5$ & $n{=}10$ & $n{=}\text{all}$ & $\Delta$ \\
            \midrule
            \multirow{4}{*}{\shortstack[l]{ManipTrans\\(Inspire)}}
              & cup   & 64.5 & 60.3 & 56.8 & 42.2 & $-34.6\%$ \\
              & spoon & 51.9 & 46.2 & 41.7 & 44.4 & $-14.5\%$ \\
              & stick & 21.1 & 11.9 &  2.0 &  1.8 & $-91.5\%$ \\
              & \emph{avg.} & \emph{55.7} & \emph{50.4} & \emph{46.1} & \emph{40.7} & \emph{$-26.9\%$} \\
            \midrule
            \multirow{4}{*}{\shortstack[l]{ManipTrans\\(WUJI)}}
              & cup   & 67.8 & 65.6 & 58.4 & 25.1 & $-63.0\%$ \\
              & spoon & 55.2 & 51.0 & 47.6 & 40.4 & $-26.8\%$ \\
              & stick & 17.2 &  7.1 &  2.8 &  3.1 & $-82.0\%$ \\
              & \emph{avg.} & \emph{58.5} & \emph{54.9} & \emph{49.7} & \emph{30.0} & \emph{$-48.7\%$} \\
            \midrule
            \multirow{4}{*}{\shortstack[l]{\methodname\\(Ours, WUJI)}}
              & cup   & 80.7 & 78.3 & 76.0 & 70.1 & $-13.1\%$ \\
              & spoon & 58.1 & 63.3 & 61.7 & 67.1 & $+15.5\%$ \\
              & stick & 43.3 & 43.7 & 39.6 & 46.7 & $+7.9\%$ \\
              & \emph{avg.} & \textbf{\emph{67.5}} & \textbf{\emph{68.9}} & \textbf{\emph{66.8}} & \textbf{\emph{67.2}} & \textbf{\emph{$-0.4\%$}} \\
            \bottomrule
        \end{tabular*}
\end{minipage}\hfill
\begin{minipage}[t]{0.45\linewidth}
    \centering
    \scriptsize
    \setlength{\tabcolsep}{3pt}
    \caption{\textbf{Real-world results.} Stage success for $30$ references per object, reported as best of up to three attempts and per-attempt rates including retries.}
    \label{tab:real_world_results}
    \begin{tabular*}{\linewidth}{@{\extracolsep{\fill}}lccc@{}}
        \toprule
        Object & Pickup & Grasp pose & Place \\
        \midrule
        \multicolumn{4}{l}{\emph{Best of up to three}} \\
            Hammer & $22/30$~($73.3\%$) & $17/30$~($56.7\%$) & $15/30$~($50.0\%$) \\
            Marker pen & $21/30$~($70.0\%$) & $13/30$~($43.3\%$) & $10/30$~($33.3\%$) \\
        \midrule
        \multicolumn{4}{l}{\emph{Per attempt}} \\
            Hammer & $31/83$~($37.3\%$) & $19/83$~($22.9\%$) & $15/83$~($18.1\%$) \\
            Marker pen & $28/86$~($32.6\%$) & $14/86$~($16.3\%$) & $10/86$~($11.6\%$) \\
        \bottomrule
    \end{tabular*}
\end{minipage}
    \par\vspace{-14pt}
\end{table*}

\subsection{Scaling Multi-Trajectory Fitting on OakInk2}
\label{sec:exp_scaling}

\paragraph{Setup.}
We compare \methodname to ManipTrans~\cite{li2025maniptrans} on cup ($93$ trajectories), spoon ($98$), and stick ($13$), the three OakInk2 objects with the most recorded trajectories.
For each object and each $n \in \{1, 5, 10, \text{all}\}$, we partition the trajectories into groups of size $n$ and train one policy per group. Each policy is evaluated \emph{only on its training group}, and we report mean Trajectory Completion over trajectories, testing multi-trajectory fitting capability.
We select ManipTrans as an established OakInk2 baseline with a released dataset-processing pipeline and native support for varying the number of references per policy. The $n{=}1$ setting recovers its standard per-trajectory setup. We test its default Inspire hand and a re-tuned WUJI variant.
We also attempted to train HOT~\cite{wang2025learning} directly on OakInk2, retaining its default hand configuration and replacing its training data with OakInk2 trajectories. This adaptation did not consistently learn successful tracking, so we do not include it as a quantitative baseline.
\methodname uses the same retargeter training pipeline, with the reference tactile input and tactile-guided rewards removed since OakInk2 provides no glove signal.

Across all training-group sizes, we train each \methodname policy for $24$ hours ($614$\,M environment steps) and each ManipTrans policy for $8$ hours ($380$\,M environment steps), on L40S hardware. These fixed method-specific budgets were selected based on observed learning plateaus and are not compute-matched.

\paragraph{Results.}
Table~\ref{tab:exp2_oakink2} reports completion on all three objects, the per-object relative change $\Delta$ from $n{=}1$ to $n{=}\text{all}$, and the trajectory-weighted average across objects.
Two trends stand out.
First, \emph{\methodname's average completion is essentially flat across $n$}, whereas both ManipTrans variants degrade sharply.
This supports our pipeline's ability to fit many reference trajectories with one policy, whereas the tested ManipTrans variants lose performance as the training group grows.
One exception for our method is cup, which steadily declines as $n$ increases, because in a few of its trajectories the object immediately drops at the initial hand pose---a case where a dedicated single-trajectory policy can more easily specialize to a quick grasp than one shared policy.
Second, \emph{\methodname achieves the highest average completion at every $n$, including $n{=}1$}, showing that the retargeting pipeline is effective on these additional hand--object tracking tasks.
Our policy achieves these results with a smaller model (1.9\,M parameters vs.\ ManipTrans's 2.3\,M).

Although \methodname uses a larger per-policy budget, sharing policies reduces aggregate training time relative to per-trajectory fitting. Across the $204$ OakInk2 trajectories, training one \methodname policy per object requires approximately $3\times24=72$ aggregate policy-training hours, compared with $204\times8=1{,}632$ hours for per-trajectory ManipTrans.

\vspace{-1mm}
\subsection{Tactile-Free Real-World Deployment}
\label{sec:exp_real_world}
\label{app:real_world_results}
\vspace{-1mm}

\paragraph{Setup.}
We evaluate the distilled student controller $\pi_{\text{ctrl}}$ on the physical xArm7--WUJI setup with a wrist-mounted depth camera. For each object, a single student is distilled from the retargeter trained on all collected reference trajectories, with no task-specific tuning at deployment. The controller follows target object trajectories using visual--proprioceptive feedback without tactile sensing. We evaluate $30$ references per object sampled from the training set, testing real-world execution rather than held-out-reference generalization.

\paragraph{Protocol.}
We score three sequential stages: \emph{Pickup}, lifting the tool off the table; \emph{Grasp pose}, reorienting it in-hand into the reference functional grasp; and \emph{Place}, completing the full reference trajectory and returning the tool to the table. Our primary result uses a best-of-up-to-three protocol: each reference is attempted up to three times and scored by the furthest stage reached across these attempts.

This protocol is motivated by initial object placement. The target trajectory specifies an initial object pose that is exact in simulation, but manually reproducing it on the real table introduces small offsets that can cause pickup failure. Allowing a small number of attempts accommodates variability in manual initialization while keeping the controller unchanged. Best-of-three therefore measures execution with limited retries, not single-attempt reliability. For completeness, we also report stage success over individual recorded attempts, including retries, alongside the main results.

\paragraph{Results.}
Table~\ref{tab:real_world_results} shows that $76.5\%$ (hammer) and $76.3\%$ (marker pen) of failed attempts end before pickup. Once the functional grasp is reached, $78.9\%$ and $71.4\%$ complete the trajectory, respectively. Grasp acquisition is therefore the main bottleneck; initial-placement offsets are a plausible contributing factor.

Beyond the quantitative evaluation on collected references, the project website shows qualitative examples of the student following rule-generated marker-pen trajectories to write letters in simulation and the real world. These examples illustrate execution on synthesized targets.
%===============================================================================
\section{Conclusion}

We present \methodname, a tactile-guided RL framework that uses glove tactile signals to guide contact acquisition and human-like grasp formation. A single generalizable retargeter learns from trajectories of the same object and is distilled into a tactile-free controller. Experiments demonstrate the benefits of tactile guidance, held-out generalization, and real-world execution.

%===============================================================================
\bibliographystyle{IEEEtran}
\bibliography{references}

\end{document}